\documentclass[conference]{IEEEtran}
\IEEEoverridecommandlockouts

\usepackage{cite}
\usepackage{amsmath,amssymb,amsfonts}
\usepackage{algorithmic}
\usepackage{graphicx}
\usepackage{textcomp}
\usepackage{xcolor}

\def\BibTeX{{\rm B\kern-.05em{\sc i\kern-.025em b}\kern-.08em
    T\kern-.1667em\lower.7ex\hbox{E}\kern-.125emX}}
\begin{document}

\title{Improved Molecular Generation through\\ Attribute-Driven Integrative Embeddings and\\ GAN Selectivity
}

\author{\IEEEauthorblockN{Nandan Joshi}
\IEEEauthorblockA{\textit{Whiting School of Engineering EP Program} \\
\textit{Johns Hopkins University}\\
Laurel, MD USA \\
njoshi13@jh.edu}
\and
\IEEEauthorblockN{Erhan Guven}
\IEEEauthorblockA{\textit{Whiting School of Engineering EP Program} \\
\textit{Johns Hopkins University}\\
Laurel, MD USA \\
eguven2@jh.edu}
}

\maketitle

\begin{abstract}
The growing demand for molecules with tailored properties in fields such as drug discovery and chemical engineering has driven advancements in computational methods for molecular design. Machine learning based approaches for de-novo molecular generation have recently garnered significant attention. This paper introduces a transformer-based vector embedding generator combined with a modified Generative Adversarial Network (GAN) to generate molecules with desired properties. The embedding generator utilizes a novel molecular descriptor, integrating Morgan fingerprints with global molecular attributes, enabling the transformer to capture local functional groups and broader molecular characteristics. Modifying the GAN generator loss function ensures the generation of molecules with specific desired properties. The transformer achieves a reconversion accuracy of 94\% when translating molecular descriptors back to SMILES strings, validating the utility of the proposed embeddings for generative tasks. The approach is validated by generating novel odorant molecules using a labeled dataset of odorant and non-odorant compounds. With the modified range-loss function, the GAN exclusively generates odorant molecules. This work underscores the potential of combining novel vector embeddings with transformers and modified GAN architectures to accelerate the discovery of tailored molecules, offering a robust tool for diverse molecular design applications.

\end{abstract}

\begin{IEEEkeywords}
de-novo molecular generation, generative adversarial networks, molecular descriptor design, transformer embeddings
\end{IEEEkeywords}

\section{Introduction}
\subsection{Background}
There is an ever-increasing demand for molecules with tailored properties for various applications such as drug discovery, semiconductors, and chemical engineering. Traditionally, identifying such molecules involves high-throughput screening of existing databases \cite{b1,b2}, a process that can be slow and expensive. Given the vastness of chemical space (it is estimated that between $10^{33}$ to $10^{60}$ chemicals could be synthesized \cite{b3,b4}), any large database still contains only a fraction of the possible design space of the compounds. Computational de-novo molecular generation offers an alternative approach for exploring the vast chemical space by selectively proposing new molecular candidates likely to satisfy the desired properties.

Recent advances in deep learning have accelerated computational molecular generation, utilizing machine learning (ML) techniques for de-novo molecular design. Notable approaches include Variational Auto-Encoders (VAEs) with Recurrent Neural Networks (RNNs) \cite{b5}, molecular graphs with Generative Adversarial Networks (GANs) \cite{b6}, reinforcement learning with RNNs \cite{b7}, and genetic algorithms with RNN-based autoencoders \cite{b8}. Most of these models use the SMILES (Simplified Molecular Input Line Entry System) \cite{b9} format for molecule representation.

While earlier works \cite{b5,b7} used sequential models like RNNs, GANs have generally outperformed VAEs and RNNs in generating novel samples. However, the standard,  vanilla GAN model lacks selectivity (the ability to generate only molecules with the desired properties selectively).  This challenge has been addressed in some works through modified training loops, such as using genetic algorithms \cite{b8}, though these involve manual curation of input data. The present work uses a modified GAN loss function, which penalizes non-compliant samples and ensures targeted output generation, as will be discussed in a subsequent section. 

Besides the GAN architecture, the vector embeddings used for the generative models play a key role in molecular generation. In recent years, there has been a growing interest in leveraging transformers for machine learning-based tasks in chemistry. Introduced in Vaswani et al. \cite{b10}, transformers have successfully handled sequential data, overcoming many limitations associated with traditional sequence-to-sequence models like RNNs by using an attention mechanism that prioritizes the importance of each token for a decoding step. 

Inspired by Large Language Models (LLMs), developed for natural language processing, various pre-trained transformer models have been developed for chemistry\cite{b11,b12, b13,b14}. These models are pre-trained on large datasets of chemical SMILES strings using Masked Language Modeling (MLM) \cite{b15}  and generate vector embeddings that can be used for property predictions. Since they are primarily focused on creating a vector encoding using MLM, none of these models are generative in nature. Only recently have some research efforts explored using transformers trained on SMILES for generative tasks \cite{b16,b17}. A new generative vector embedding has been developed in the present work and will be discussed later. 

Due to its compactness, the SMILES format has been the preferred molecular representation for most of these models discussed previously (both transformer models and RNNs). SMILES uses various grammatical rules to represent complicated 3D molecular structures as a linear string. However, using a SMILES string directly to generate embeddings for molecular generation tasks has certain limitations.

The structure of a molecule plays a crucial role in its chemical properties. Its functional groups and neighborhood interactions drive the chemical properties of a molecule. SMILES notation simplifies molecular structures such as rings into a linear string, inevitably losing information about functional groups and their spatial relationships. Unless a SMILES string is converted into a molecular graph, essential details about atom neighborhoods and functional groups are not explicitly represented. Without this explicit representation, a transformer trained on raw SMILES strings may fail to capture these critical features, learning the syntactical rules of the SMILES grammar rather than identifying the key atom groups that influence the molecule's properties\cite{b6}.

Furthermore, factors such as aromaticity, resonance, and electronegativity, which play essential roles in a molecule’s chemical properties, are not captured in the SMILES format. Consequently, SMILES may not be the most suitable representation for molecular property prediction and associated generative tasks. Some studies on property prediction choose not to use a raw SMILES strings directly, instead relying on circular fingerprints that capture more comprehensive neighborhood information \cite{b18,b19}. As a result, there is a need for a different molecular input vector that captures these key attributes and can be used for generating vector embeddings, which is the motivation for this work. 

\subsection {Objective and Contribution}
This work introduces a generative framework that leverages transformer embeddings derived from a novel molecular descriptor to generate novel molecules. There are three key novel components to this work. 
\begin {enumerate}
\item A new molecular descriptor, which combines molecular fingerprints with global molecular attributes, is introduced as an alternative to using the SMILES format directly
\item The use of a transformer encoder-decoder to create generative vector embeddings
\item Modified GAN loss function to ensure selectivity in output generation. 
\end{enumerate}

A test case is presented in which the generated embeddings are used to produce odorant molecules. The effectiveness of the embedding generator will be assessed by evaluating the accuracy of SMILES string conversions. The performance of the modified GAN loss function will be compared to that of a vanilla GAN by analyzing the proportion of odorants generated. Additional evaluation will include metrics such as novelty, uniqueness, diversity, and the validity of the generated SMILES strings.

The paper is outlined as follows. Section 2.A introduces the new molecular descriptor. Section 2.B illustrates the transformer architecture with the embedding generator. Section 2.C describes the modified GAN function. Section 3 enlists all the model details, their hyperparameters, and datasets. Section 4 discusses the results of the experiments. 

\section{Proposed Approach: Modified GAN with Novel Generative Embeddings}
\subsection{Novel Molecular Descriptor}
This work introduces a novel molecular descriptor to address the limitations of SMILES strings. Specifically, the Morgan fingerprints \cite{b20} of a molecule and other important global molecular attributes are used as an input to the transformer to generate vector embeddings. 
\begin{enumerate}
    \item Morgan Fingerprints: Morgan fingerprints \cite{b20} are binary bit vectors that capture the neighborhoods of all atoms in a molecule, thereby encoding essential features like functional groups and their interactions with other groups. They are highly effective for property prediction and molecular similarity searching tasks. Using them instead of SMILES makes the transformer learns all the molecular neighborhoods and functional groups that drive the chemical properties, making them more suitable for de-novo molecular generation based on predicted properties. A previous study \cite{b21} found that transformers can be used to reconstruct associated SMILES strings from Morgan fingerprints with a high accuracy of 93\%, indicating minimal information loss with these fingerprints. 
    \item Global Molecular Attributes: Properties such as  molecular weight, Information Content (IpC), and aromaticity play a major role in a molecule's chemical behavior. The cheminformatics package, RDKit \cite{b22}, generates about 200 such attributes from a given SMILES string. which, if combined with Morgan fingerprints, would provide richer information to a transformer for property prediction. \end{enumerate}
    After removing redundant features, 30 of the RDKit attributes and Morgan fingerprints are to form an input vector for the transformer, capturing key structural features and important molecular properties.

\subsection{Embedding Generator using Transformers}
\begin{figure}[!htb]
    \centering
    \includegraphics[scale=0.50]{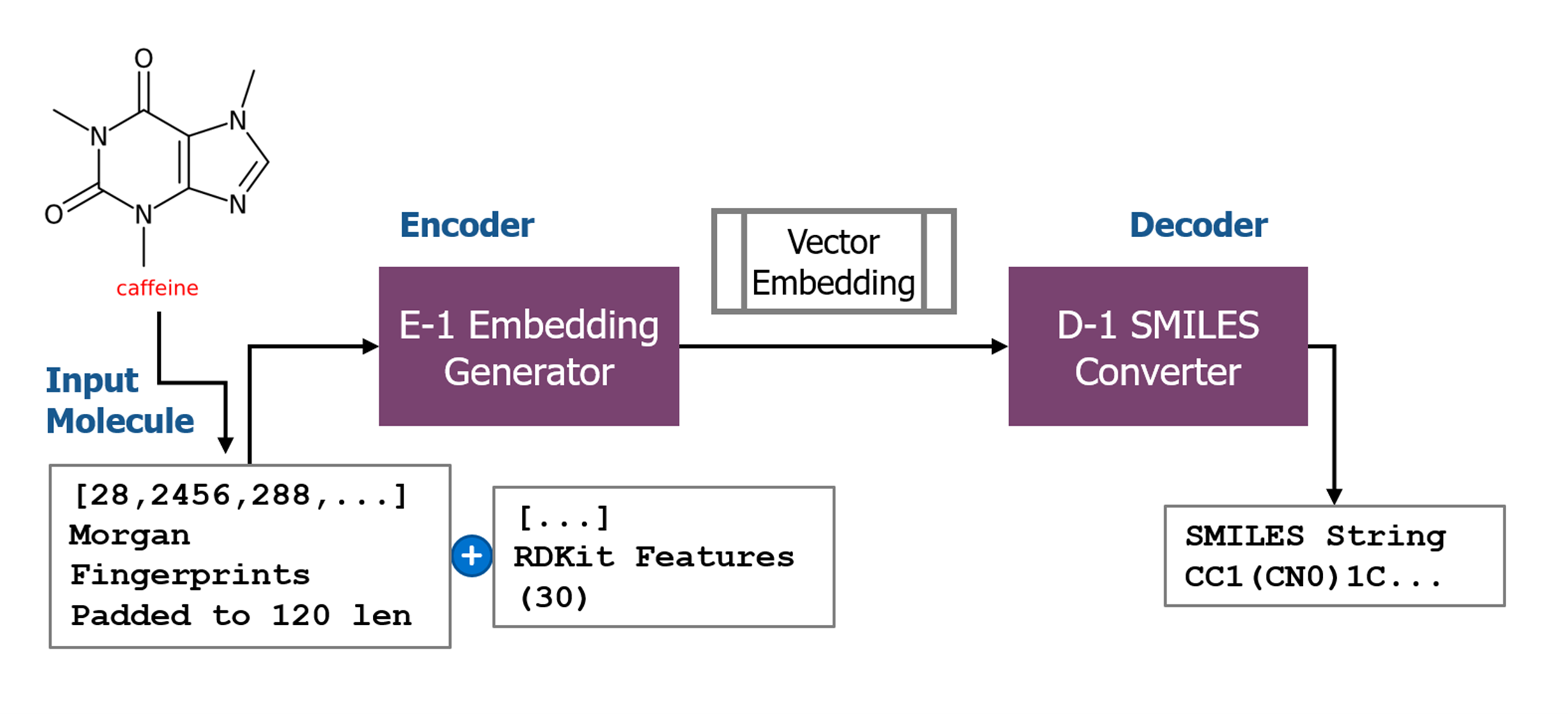}
    \caption{Transformer Architecture}
    \label{fig1}
\end{figure}

Fig.~\ref{fig1} outlines the transformer architecture used for generating the embeddings. The input vector (molecular descriptor) is constructed as described in the previous section. The transformer is trained to generate the SMILES string of a molecule using this molecular descriptor. A direct token-to-token comparison of the transformer's output with the target SMILES is used and trained to minimize the negative log-likelihood loss. It is pre-trained using a large unlabeled dataset of SMILES strings, which will be discussed in Section 4.

The transformer is divided into two parts. The embedding generator (E-1) is the encoder of the transformer. It applies multiple attention layers to  the source sequence (Morgan fingerprints + 30 RDkit features) to generate a vector embedding. The SMILES generator (D-1) is a transformer decoder that recreates the SMILES sequence from the vector embedding. After the transformer is pre-trained using a large dataset, the vector embeddings created by E-1 can be used for downstream tasks like property prediction and molecular generation. The advantage of training the decoder to output SMILES strings is that it allows for reliable translation of any new embedding back into a valid SMILES string, enabling generative tasks.

\subsection{Modified GAN for Molecular Generation}
GANs\cite{b23} are a class of models that have been successfully applied to generative tasks, including molecular generation. GANs consist of two networks: a generator, which creates new data samples, and a discriminator, which distinguishes real samples from fake. The generator and discriminator are trained together, trying to outcompete each other and thus improving iteratively. A traditional GAN comes with many challenges, such as training instability and mode collapse. To address these issues, Wasserstein GANs (WGANs)\cite{b24} use a different objective function based on Wasserstein distance (Earth Mover’s Distance) , which provides a more stable and reliable optimization process.

\begin{figure}[!htb]
    \centering
    \includegraphics[scale=0.35]{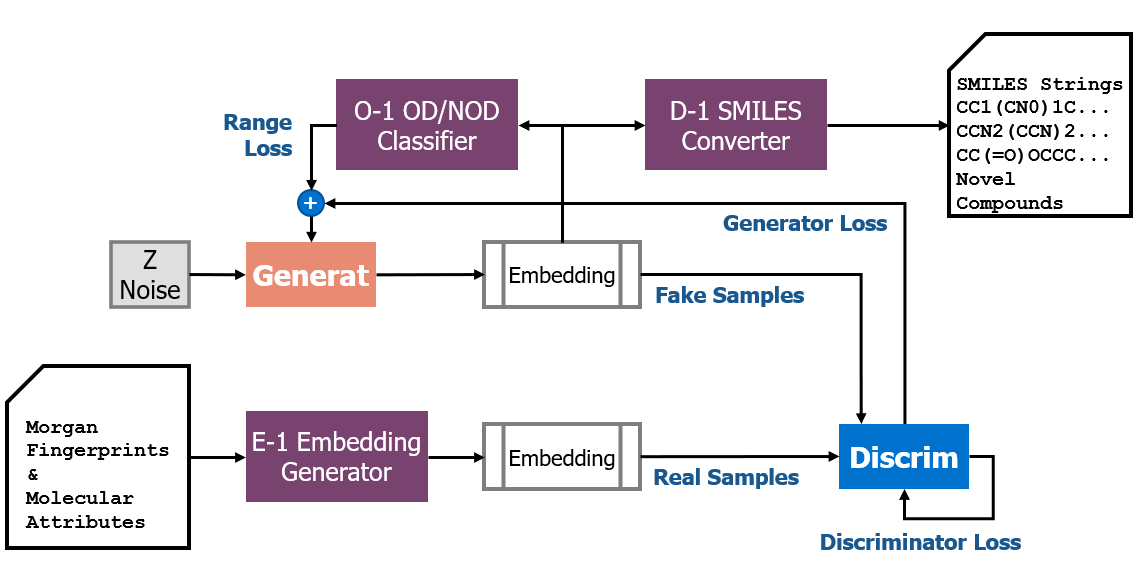}
    \caption{Range-constrained GAN Architecture}
    \label{fig2}
\end{figure}

A WGAN is utilized in this work to generate novel molecules, along with a modified loss function for the generator. Fig.~\ref{fig2} displays the GAN architecture. The embeddings generated by the embedding generator of the transformer (E-1) are used as "real samples" to train the discriminator. The generator generates a vector of the same dimensions as the embedding from a noise vector. The decoder (D-1) translates the newly generated vector embeddings into valid SMILES strings. In addition to the standard generator loss, an additional range-loss is introduced (as will be discussed in Subsection 1), and added to the generator. The newly generated vector embeddings are passed to an OD/NOD classifier (O-1), and the range-loss reflects the proportion of non-odorant samples. By minimizing this loss, the generator is thus effectively trained to generate only odorants.

\subsubsection {Range-Loss: Modified Loss Function to Generate Molecules with Specific Traits}

The loss function for the generator is modified to generate more odorant molecules selectively. This approach is adapted from Nobari et al. \cite{b25}, where this loss function was first proposed. The Generator objective is modified as Equation 1

\begin{equation}
    \underset{G}{\min}(-1 \cdot \mathbb{E}_{z\sim P_{z}}[D(G(z)] + \lambda_1 \cdot \mathcal{L}_{rng})
\end{equation}

The term $\mathbb{E}_{z\sim P_{z}}[D(G(z)]$ represents the standard Wasserstein generator loss function \cite{b24}, whereas $L_{rng}$ is the added range-loss to ensure selective output generation. 

Since $L_{rng}$ provides selectivity to the GAN, it should penalize any samples lying outside the given property range. 
The form of this equation can be defined as in Eq 2

\begin{equation}   
    \mathcal{L}_{rng}=-\frac{\sum\limits_{i=1}^N 1_{(y_i - y_{i,lb}) \cdot (y_i - y_{i,ub}) \geq 0 } \log p(x_i | [y_{i,lb}, y_{i, ub}])}{\sum\limits_{i=1}^N 1_{(y_i - y_{i,lb}) \cdot (y_i - y_{i,ub}) \geq 0 }}
\end{equation}

The term  ${\sum\limits_{i=1}^N 1_{(y_{i} - y_{i,lb}) \cdot (y_{i} - y_{i,ub}) \geq 0 } }$ is zero for all samples lying within the lower and upper bounds ($y_{i,lb} < y < y_{i,ub}$). This term enforces that only the samples non-compliant samples contribute to an additional loss, with more non-compliant samples leading to a larger loss function value.  

The actual value of the loss is calculated by the probability of condition satisfaction, $p(x_i | [y_{i,lb}, y_{i, ub}]))$. This function should have a zero gradient for samples within the satisfaction range, and also have a gradient that decreases for samples lying closer to the condition satisfaction range for stable training. The authors of \cite{b24} proposed a form that is analogous to the negative log likelihood (NLL) (Eq3) , and thus satisfies both the requirements

\begin{equation}
    p(x_i | [y_{i,lb}, y_{i, ub}]) = \frac{1}{1 + \exp^{\phi (\mathbb{E}(x) - y_{lb})}} - \frac{1}{1 + \exp^{\phi (\mathbb{E}(x) - y_{ub})}}
\end{equation}

While this range-loss function has been developed for continuous output in a regression problem, the odorant classification considered in this work is a binary classification. So, the binary cross-entropy (BCE) of the odorant classifier output is considered for the range-loss calculation, with its intended range lying between 0.5 and 1.\par

\section {Model Hyperparameters and Datasets}
\subsection {Transformer (Embedding Generator)}
The general architecture of the transformer follows the design outlined in the original transformer paper by Vaswani et al. \cite{b10}, with the same hyperparameters. The source sequence is padded to a length of 150, while the target sequence (SMILES strings) is padded to a length of 74. The encoder and the decoder use six attention layers with eight heads and a dimension of 512. Batch normalization and a dropout layer (0.1) are used before each attention layer. Adam's optimizer is used with $\beta_1$ and $\beta_2$ of 0.1 and 0.99, respectively. The vector embeddings produced by the transformer’s encoder have a dimension of 150x512 and are then used as inputs for downstream models, namely the GAN and the odorant classifier.

\subsection{GAN Model Parameters}
The discriminator consists of five 1D convolutional layers of kernel size 3 and reduces the 150x512 encoding vector to a 150x16 vector, which is sent through a final fully connected layer to produce a single discriminator output.

The generator takes a noise vector of 1x512 dimensions as an input. It is composed of 5 1D convolutional layers of kernel size 3, where the channel number is quadrupled for each layer. Thus, the 1x512 noise vector is converted to a vector of dimension of 1024x512, which is finally sent through a fully-connected layer to obtain the generator output of size 150x512. This generator output represents the vector embedding of a novel molecule. It is decoded into a SMILES string by sending it to the decoder D1.

The GAN training loop uses a learning rate of 0.001 with a Stochastic Gradient Descent (SGD) optimizer. To avoid mode collapse, the weights are clipped to 0.1.For the additional Range-loss, a hyperparameter of $\lambda_1 = 10$ is used. The exponential factor $\phi$ is set to 10. 

The GAN progressively starts generating more realistic molecules as it progresses through training.
\subsection{Odorant Classifier}
An odorant classifier (O-1) is trained on the transformer’s vector embeddings (E1). The odorant classifier uses the embedding (150x512) as an input and employs convolutional neural networks. A total of 7 Convolutional Neural Networks (CNN) layers are used with a kernel size of 5x5, stride of 2x2, and a padding of 2x2. The first layer outputs 16 channels from one channel. The next three layers double the output channels each, and the last three layers do not change the channel number. The final output (128x6x4) is sent through a final fully-connected layer to produce the classification output. Batch normalization and dropout layers (dropout probability = 0.85) are used before each CNN layer. The classifier employs a SGD optimizer with a learning rate (LR) of 0.001.

\subsection{Datasets}
The ZINC database\cite{b26}, containing approximately 13 million commercially available drug-like compounds, is used for pre-training the transformer.  20\% of this dataset is used (2.6M) for pretaining the model. The data is split into an 80\%-20\% train-test split.

For training the odorant/non-odorant classifier (O-1), labeled data from the DeepOlf model \cite{b27},labeled data from the DeepOlf model, consisting of about 3,800 SMILES strings, is used. Approximately 80\% of the data is classified as odorants, with the remaining as non-odorants, thus making it unbalanced. This dataset is also split 80\%-20\% for training and testing, maintaining the original class balance. The database is broad, as it includes odorants without specifying which receptors detect them, presenting a challenge for molecular generation due to the diversity of receptor interactions.

For training the Range-GAN, the entire ZINC database is used. For experimental trials, it is also compared with a vanilla GAN trained on odorant-only compounds. To obtain the odorant-only dataset, the ZINC 2.6M dataset is filtered using the OD/NOD classifier to identify only the odorants, and a reduced dataset of 13555 molecules is obtained.

\section{ Experimental Results}
\subsection {Validation of the transformer architecture}
Since the transformer is used for generative tasks, it has to be validated by measuring the conversion accuracy of SMILES strings from the embeddings.Testing with a 20\% validation set from the 2.6M ZINC dataset showed that the transformer achieved a 94\% conversion accuracy.  Ucak et al. \cite{b21} demonstrated that a transformer decoder using only Morgan fingerprints reached 93.1\% conversion accuracy. The slight improvement in accuracy observed here, with a smaller dataset, suggests that adding RDKit features to the Morgan fingerprints enhances both convergence and accuracy. This high reconversion accuracy supports the transformer's capability to reliably generate valid SMILES strings for novel compounds using the proposed molecular descriptor, which is critical for validating its effectiveness in generative tasks like GANs.

\subsection{Odorant Classifier Performance}
The Odorant Classifier (O-1) is first validated before it is used for generating GAN outputs. As mentioned in a previous section, the DeepOlf dataset is split with an 80-20\% train-test split. After training, the results of the classifier on the validation (testing) set are tabulated in Table I. 

\begin{table}[!htb]
    \centering
    \begin{tabular}{|c | c | c | c |}
         \hline
          Accuracy & Precision & Recall & F1 Score  \\
         \hline
          95.0\% & 94.6 \% & 93.8 \% & 94.2\% \\
         \hline
    \end{tabular}
    \caption{Odorant Classifier Performance}
    \label{tab2}
\end{table}

Since this classifier would be used for proposing new odorants, the precision score is the most important metric since false positives would lead to wasteful suggestions. The model performs fairly well on this metric, with a near 95\% precision score. This aligns well with the baseline DeepOlf \cite{b27} model. The good accuracy reinforces confidence in using the generated embeddings for the purpose of property prediction. 

\subsection {Experimental Results with GAN}

To test the utility of Range-loss, the outputs generated with it are compared with the outputs generated with a vanilla GAN trained on an odorant-only dataset. Table II shows that Range-Loss is very powerful when generating  odorants exclusively. On the contrary, the vanilla GAN trained only with odorants  misses out on 48\% of the molecules it generates. 

\begin{table}[!htb]
    \centering
    \begin{tabular}{|c | c |}
         \hline
         Proportion of odorants produced & \% Odorants  \\
         \hline
         
          GAN with Range-Loss&99.2\%\\
         \hline
         Vanilla GAN trained only on odorants & 52\%\\
         \hline
    \end{tabular}
    \caption{Proportion of odorant molecules produced}
    \label{tab2}
\end{table}

A key challenge with GANs is balancing specificity with diversity and uniqueness. Table III shows how the two GANs work on various metrics. 

\begin{table}[!htb]
    \centering
    \begin{tabular}{|c | c c c c|}
         \hline
          GAN type & Uniqueness & Validity & Novelty & Internal Div.  \\
         \hline         
          Vanilla GAN & 99\% & 67\% & 100\% & 0.81\\         
         \hline
          Range-Loss & 99\% & 62\% & 100\% & 0.77\\   
          \hline
    \end{tabular}
    \caption{GAN performance metrics}
    \label{tab1}
\end{table}

There is baseline data from \cite{b8} that shows the GAN performance with a SMILES sequence and RNN-based embeddings. This baseline model had a validity of 30.2\%, far lower than the proposed approach. The increased validity underscores the assertion that introducing more descriptive embeddings helps generate molecules that map to realistic SMILES formulae. 

In this context, internal diversity was measured as the mean Tanimoto similarity between the molecular fingerprints of all molecules in the dataset.

Table III shows the performance to be comparable between the two models, indicating that range loss does not cause degradation in the GAN performance. A slight reduction in diversity with range-loss was anticipated due to the specificity of outputs.

\begin{figure}[!htb]
    \centering
    \includegraphics[scale = 0.33]{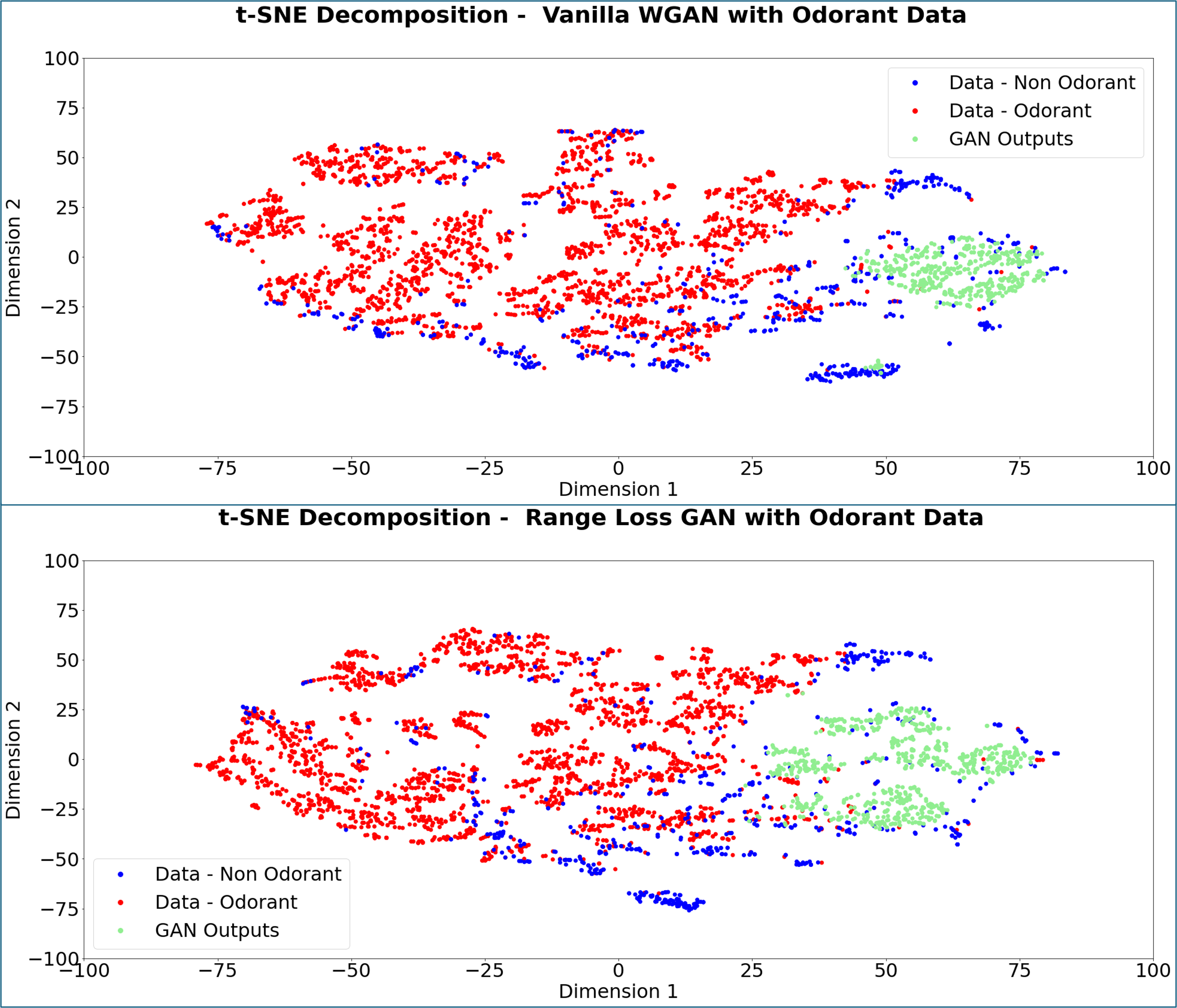}
    \caption{Values of Range-Loss GAN with 15k elements of ZINC Dataset - t-SNE decomposition}
    \label{fig5}
\end{figure}

 Figure ~\ref{fig5} offers further insight into the performance comparison, with t-SNE decomposition applied to a vector of the 30 RDKit attributes of all generated molecules. The scatter of generated outputs shows that molecules from the range-loss GAN are closer to the odorant cluster, demonstrating that the modified GAN generates more odorant-like compounds. However, the odorant datapoints are widely spaced due to the broad nature of the odorant database, which includes compounds without specificity to particular receptors. A more focused database with odorants targeting specific receptors could improve the quality of generated molecules, which will be considered for future work.

\subsection{Model Complexity and Training Time}

Some discussion is also needed regarding the model complexity and training time. The most time-consuming step during the training is training the embedding generator (E-1), and the largest model is also E-1. Trained with 2.6M SMILES, with a batch size of 32, it takes a total of 15 epochs for the model to converge (with the final epoch being the conversion rate of 94\%). The total cold-start training time on a NVIDIA  RTX A5000 GPU is about 6 days. 

\section{Conclusions and Future Work}
This work introduces a machine-learning framework for generating molecules with specific properties. It has three main novelties -  a new molecular descriptor, a generative vector embedding using transformers, and a GAN with a loss function that directs it to produce specific outputs. 

The new molecular descriptor combines a molecule's Morgan fingerprints (radius 2) with some other global molecular attributes generated using RDKit. Thus, the proposed descriptor provides a more comprehensive representation of a molecule's functional groups and broader molecular characteristics, which are the important drivers of its chemical properties.

The transformer is trained such that the vector embeddings generated by the embedding generator are decoded to reconstruct the SMILES string of the original molecule. As a result, the vector embeddings are generative in nature since any new embedding generated by a GAN can be mapped to a valid SMILES string. The transformer is pre-trained with a large unlabeled dataset of molecules from the ZINC database. Testing with a 20\% hold-out set reveals that the decoder can accurately recover 94\% of the original SMILES strings of the molecules from their vector embeddings. This result reinforces confidence in using the proposed embedding generator for generation tasks.

A modified loss function is added to the GAN generator loss. This "Range-Loss" penalizes any molecules that do not possess properties lying within a desired range. By doing so, it forces the generator to produce only those molecules which have the desired properties. A test case is considered, where the GAN produces odorant compounds. It is found that the GAN with "Range-Loss" generates almost all  odorants based on their classification. In comparison, a vanilla GAN that is trained only using odorants produces just 52\% of odorants. This shows the power of adding the range-loss to obtain the desired molecules.

The results of the GAN show a very high degree of novelty, uniqueness, and diversity. Most importantly, the experiments showed that about 62\% of the samples generated by this GAN were valid SMILES strings, showing that the embeddings are mapped to valid molecules. Compared to a baseline result \cite{b8} that showed a validity of 30.2\% using RNNs, the results of the current work display a big improvement. In addition, in comparison with vanilla GAN, adding the range-loss does not cause a significant deterioration in GAN metrics like novelty, uniqueness, and validity. 

The t-SNE decomposition showed that the odorant input space was far too broad compared to the generated outputs. One reason for this is that the odorant dataset contains a very broad range of chemicals, with very different types responding differently to different types of odorant receptors. A further improvement of this work would be to target odorants connected with specific receptors to reduce the design space and provide GAN with a more constrained problem. 

Overall, this work highlights the potential of combining informative molecular representations with neural network models to accelerate the discovery and design of molecules with tailored properties. It introduced a framework to generate molecules using these intergative embeddings. In many practical applications, there is a need for molecules tailored to not just a single property, but rather a set of desired attributes. While the framework introduced here handles only a single attribute, it has the potential to be modified further for multiobjective molecular generation, which would represent a powerful addition to this framework's capabilities.

\end{document}